\documentclass[acmsmall,screen]{acmart}
\AtBeginDocument{%
  \providecommand\BibTeX{{%
    \normalfont B\kern-0.5em{\scshape i\kern-0.25em b}\kern-0.8em\TeX}}}

\setcopyright{acmcopyright}
\copyrightyear{2021}
\acmYear{2021}
\acmDOI{00.0000/000000.00000}

\received{April 2021}

\acmJournal{TIST}
\acmArticle{39}
\acmMonth{3}

\usepackage{amssymb}
\usepackage{amsmath}
\usepackage[ruled,vlined]{algorithm2e}

\usepackage[utf8x]{inputenc}

\DeclareUnicodeCharacter{65292}{ }

\usepackage{amsthm}
\usepackage{multirow}
\usepackage{booktabs}
\usepackage{pgffor}
\usepackage{epstopdf}
\usepackage{tikz}
\usepackage{indentfirst}
\usepackage{epstopdf}
\usepackage{mathtools}

\usepackage{enumitem}
\usepackage{tabulary}
\usepackage{balance}
\usepackage{flushend}
\usepackage{subfig}
\usepackage{graphicx}

\theoremstyle{definition}

\usepackage{geometry}

\begin{document}



\title{Federated Multi-View Learning for Private Medical Data Integration and Analysis}
 

\author{Sicong Che}
\affiliation{\institution{China Agricultural University}
\city{Beijing}
\country{China}
}
\email{sicongche@gmail.com}

\author{Hao Peng}
\affiliation{\institution{Beihang University}
\city{Beijing}
\country{China}
}
\email{penghao@act.buaa.edu.cn}

\author{Lichao Sun}
\affiliation{\institution{Lehigh University}
\city{Bethlehem}
\state{PA}
\country{USA}
}
\email{lis221@lehigh.edu}

\author{Yong Chen}
\affiliation{\institution{University of Pennsylvania}
\city{Philadelphia}
\state{PA}
\country{USA}
}
\email{ychen123@pennmedicine.upenn.edu}

\author{Lifang He}
\affiliation{\institution{Lehigh University}
\city{Bethlehem}
\state{PA}
\country{USA}
}
\email{lih319@lehigh.edu}

\renewcommand{\shortauthors}{S. Che et al.}
\begin{abstract}
Along with the rapid expansion of information technology and digitalization of health data, there is an increasing concern on maintaining data privacy while garnering the benefits in medical field. Two critical challenges are identified: Firstly, medical data is naturally distributed across multiple local sites, making it difficult to collectively train machine learning models without data leakage. Secondly, in medical applications, data are often collected from different sources and views, resulting in heterogeneity and complexity that requires reconciliation. This paper aims to provide a generic Federated Multi-View Learning (FedMV) framework for multi-view data leakage prevention, which is based on different types of local data availability and enables to accommodate two types of problems: Vertical Federated Multi-View Learning (V-FedMV) and Horizontal Federated Multi-View Learning (H-FedMV). We experimented with real-world keyboard data collected from BiAffect study. The results demonstrated that the proposed FedMV approach can make full use of multi-view data in a privacy-preserving way, and both V-FedMV and H-FedMV methods perform better than their single-view and pairwise counterparts. Besides, the proposed model can be easily adapted to deal with multi-view sequential data in a federated environment, which has been modeled and experimentally studied. To the best of our knowledge, this framework is the first to consider both vertical and horizontal diversification in the multi-view setting，as well as their sequential federated learning.
\end{abstract}



\authorsaddresses{
Authors' addresses: 
S. Che, College of Information and Electrical Engineering, China Agricultural University, Beijing 100083, China; email: \path{sicongche@gmail.com}; 
H. Peng, Beijing Advanced Innovation Center for Big Data and Brain Computing, Beihang University, No. 37 Xue Yuan Road, Haidian District, Beijing, 100191, China; email: \path{penghao@act.buaa.edu.cn}; 
L. Sun, Department of Computer Science and Engineering, Lehigh University, Bethlehem, PA; email: \path{lis221@lehigh.edu};
Y. Chen, Department of Biostatistics, Epidemiology and Informatics, University of Pennsylvania, Philadelphia, PA; email: \path{ychen123@pennmedicine.upenn.edu};
L. He, Department of Computer Science and Engineering, Lehigh University, Bethlehem, PA; email: \path{lih319@lehigh.edu}.
}

\maketitle

\vfill\eject
\section{Introduction}\label{sec:introduction}
With the recent advance of technology, medical treatment has gradually become digitized, and a large and heterogeneous amount of medical data (e.g., EHR, claims, laboratory tests, imaging and genomics) has been accumulated in medical institutions. The availability of these data offers many potential benefits for healthcare. It not only facilitates sharing information in care-related activities, but also reduces medical errors and service time. Moreover, data sharing in the multi-institutional and large-scale collaboration context can effectively improve the generalizability of research, accelerate progress, enhance collaborations among institutions, and lead to new discoveries from data pooled from multiple sources \cite{lu2020web}.
For these reasons, substantial efforts have been made in the past to improve the medical information systems enabling the collection and processing of huge amounts of data to work toward a multi-institutional collaborative setting. For example, District Health Information Software 2 (DHIS2) \cite{dehnavieh2019district} is an open-source, web-based Health Management Information System (HMIS) widely used by countries for national-level aggregation of medical data. With the broad adoption of HMISs, security and privacy become critical \cite{singer1993impact}, as medical data are highly sensitive to the patients.

In the last two decades, great efforts have been made to facilitate the availability and integrity of medical data while protecting confidentiality and privacy in medical information systems \cite{barrows1996privacy,barach2000reporting}. Most of the research is centered around data de-identification \cite{sweeney2002k, wellner2007rapidly, el2008heuristics, el2009globally} and data anonymization \cite{deutsch2005privacy, xiao2006anatomy, machanavajjhala2007diversity, li2007t}, which removes the identifiable information from the published medical data to prevent an adversary from reasoning about the privacy of the patients. However, as pointed out in \cite{li2011new}, published medical data is not the only source that the adversaries can count on: with a large amount of information that people voluntarily share on the Web, sophisticated attacks that join disparate information pieces from multiple sources against medical data privacy become practical. 
As a result, many medical institutions are unwilling to share their data, as sharing may cause sensitive information to be leaked to researchers, other institutions, and unauthorized users. 
Even in multi-institutional collaborative research, they are expected to perform integrated analysis without leaving their medical data outside institutions. 
Therefore, security and privacy become an obstacle and challenge for data integration in medical research, especially when the data are required to be shared for secondary use.

However, machine learning requires a huge amount of data for better performance, and the current circumstances put deploying or developing medical AI applications in an extremely difficult situation. In order to address the data limitation and isolation issues, great progress has been made in the development of secure machine learning frameworks in recent years. A popular approach is the use of federated learning (FL) to support collaborative and distributed learning processes, which enables training machine learning models over remote devices or siloed data centers, such as mobile phones or hospitals, while keeping data localized \cite{mcmahan2017communication}.
For example, FL can be used to process EHR data distributed in multiple hospitals by sharing the local model instead of the patients' data to prevent the raw data leakage~\cite{boughorbel2019federated, huang2019patient, lee2018privacy, brisimi2018federated}.

In the literature, many studies have explored the variants of FL schemes to support complicated tasks in real life. For example, Yang \emph{et al.} \cite{yang2019federated} introduced a comprehensive federated learning framework that contains vertical federated learning (VFL) and horizontal federated learning (HFL) based on different local data availability. Specifically, VFL deals with the case that the datasets share the same sample ID space but their feature spaces are different, while HFL deals with the case that the datasets share the same feature space but are different in sample ID spaces. 
However, the previous works mainly focus on the difficulties in configuration of FL because of the data distributions but less consider the data complexity. In medical applications, many datasets are complex, heterogeneous and often collected from different instruments and/or measures, known as ``multi-view'' data \cite{xu2013survey}. For example, electronic health records (EHRs) contain different types of patient-level variables, such as demographics, diagnoses, problem lists, medications, vital signs, and laboratory data.
The mobile keyboard data are composed of various sensor data and keystroke records. It has been shown that multi-view learning can get better performance than the single view counterpart, especially when the strengths of one view complement the weaknesses of the other. Nevertheless, federated learning from multi-view data is still in its infancy, as most of the current solutions can only handle the single-view data. Moreover, existing works are only designed for a static setting without considering the temporal information of the data (e.g., EHRs).


Motivated by the aforementioned problems, in this paper we focus on multi-view learning tasks and associated problems of federated learning. Specifically, we first present a general Federated Multi-View Learning (FedMV) framework, which encompasses most of the existing
multi-view fusion schemes and provides freedom to create their FL counterparts. Then, based on different types of local data availability, i.e., horizontal and vertical as shown in Fig. \ref{fig:mvdata_example}, we develop two multi-view federated algorithms: V-FedMV and H-FedMV. Both V-FedMV and H-FedMV approaches can perform the same tasks, but for different multi-view data distribution. In the V-FedMV approach, each client owns a single-view data, but they have common participants. In the H-FedMV approach, every client owns multi-view data equal to a subset of the overall data. Third, we investigate how multi-view sequential data can be arranged in the setting of sequential federated learning and present a sequential modelling strategy (S-FedMV) to consider the temporal information of the multi-view data. These methods together provide flexible and effective tools to support multi-institutional collaborations for multi-view data mining while solving the privacy and security challenges for data sharing in medical research.

Our contributions can be summarized as follows:
\begin{itemize}
    \item To the best of our knowledge, we are the first group that aims to propose a systematically solution for federated multi-view learning.
    \item We have considered two types of distributed multi-view data and propose V-FedMV and H-FedMV to deal with these two cases separately.
    \item This is the first work to consider multi-view sequential data in the federated learning setting, and S-FedMV method is developed in this regard.
    \item Based on the experimental results, all three proposed methods (V-FedMV, H-FedMV, and S-FedMV) can make full use of multi-view data and are more effective than local training.
    
    
\end{itemize}

The rest of this paper is organized as follows. Section \ref{sec:2problem_definition} describes the problem definition. Section \ref{sec:mvl} introduces the multi-view learning method and optimization. The V-FedMV and H-FedMV methods are presented in Section \ref{sec:v_fedmv} and Section \ref{sec:h_fedmv}, respectively.
Then in Section \ref{sec:sequential}, we show how to adapt the model to deal with sequential data. In Section \ref{sec:privacy}, we briefly discuss the privacy issue. The dataset, experiments and results are presented in Section \ref{sec:exp}. Section \ref{sec:related} discusses the related work, followed by the conclusion in Section \ref{sec:conclusion}.

\section{PROBLEM DEFINITION} \label{sec:2problem_definition}
In this section, we state the problem of federated multi-view learning task. Without loss of generality, we consider the classification problem.
Assume that we have a distributed multi-view dataset available with $N$ instances from $K$ views: let $\mathbf{X}_k\in \mathbb{R}^{N\times d_k}$ denote the data matrix in the $k$-th view, where the $i$-th row $\mathbf{x}_{ik} \in \mathbb{R}^{d_{k}}$ is the feature descriptor of the $i$-th instance in the $k$-th view. Suppose these $N$ instances are sampled from $C$ classes and denote $\mathbf{Y}= [\mathbf{y}_1, \cdots, \mathbf{y}_N]^T \in \{0, 1\}^{N \times C}$, where $\mathbf{y}_i \in \{0, 1\}^{C \times 1}$ is the one-hot label indicator vector of the $i$-th instance. 

The key challenge to the adoption of distributed data is the security and privacy of highly sensitive data. For example, in many situations, medical data cannot leave the institutions, which limits the usefulness of these data to perform analytics in the data aggregation process. In order to solve this issue, federated learning has emerged as a promising technique for distributing machine learning (ML) model training, which performs local model training, and upload model parameters for global aggregation; thus, it enables collaborative model training while preserving each participant's privacy, which is particularly beneficial to the medical field.


Based on the local data availability, we suppose that there are two types of federated multi-view learning: horizontal and vertical. Fig. \ref{fig:mvdata_example} shows the example of vertical and horizontal multi-view scenarios. Our goal is to leverage all available data without sharing data between institutions for model learning in these respective scenarios, by distributing the model-training to the data-owners and aggregating their results. The problems are formally defined as follows: 



\begin{figure}[t]
\centering
\subfloat[Vertical]{
\label{fig:Struc_V}
\includegraphics[width=6cm]{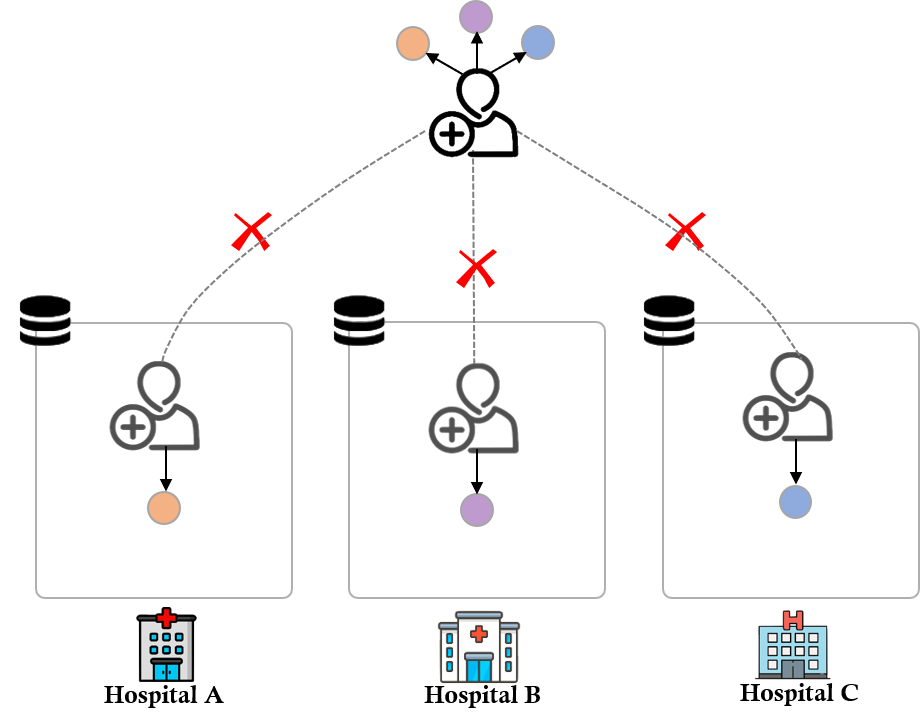}}
\subfloat[Horizontal]{
\label{fig:Struc_H}
\includegraphics[width=6cm]{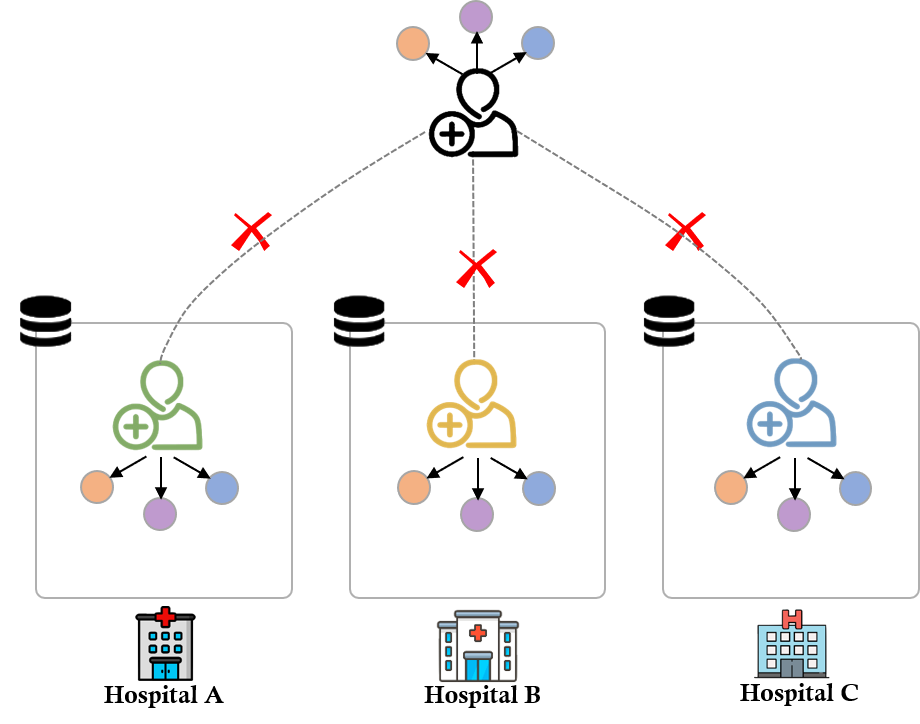}}
\caption{Two types of federated learning in multi-view scenarios.}
\label{fig:mvdata_example}
\end{figure}

\begin{itemize}
\item \textbf{Vertical Federated Multi-view Learning}: It is supposed that each client shares the same ID space but has different single-view data. For example, several hospitals want to build a model to diagnose the bipolar affective disorder, they have some patients who have bipolar affective disorder in common because these patients have been to all these hospitals.
However, each of them only has a single-view data collected from their own device. Our goal is to generate more accurate and robust classification results, by collectively training multiple single-view models through federated learning, so as to exploit vertical-level view information across multiple decentralized sites holding local private data samples without exchanging.

\item \textbf{Horizontal Federated Multi-view Learning}: It is supposed that each client owns multi-view data but shares different ID spaces, and each client's data can be seen as a subset of the overall data. For example, each hospital has multiple devices and can obtain multiple views. Besides, the types of multiple views across different hospitals are the same. However, these hospitals have few or no patients in common. Our goal is to incorporate all samples in the model training to make the result more promising, by collaboratively training multiple multi-view models with the benefit of the federated learning framework. 
\end{itemize}

\section{MULTI-VIEW LEARNING} \label{sec:mvl}
To deal with the federated multi-view learning problems in the vertical and horizontal situations, we start with a basic multi-view learning method to fuse multi-view data in this section, and then show how the formulas are modified to accommodate vertical and horizontal cases in the following two sections. Our approach is inspired by \cite{feng2020multi}, which extended the idea of multi-view learning to study the multi-class VFL problems involving multiple parties. 

\subsection{Objective Function}
Let's define $\mathbf{X}=\{\mathbf{X}_1,\mathbf{X}_2,\cdots,\mathbf{X}_K\}$ as the data matrix with K views, where $\mathbf{X}_k\in \mathbb{R}^{N\times d_k}$ is the $k$-th view data, $k=1,\cdots,K$. The objective function of multi-view learning (MVL) can be formally written as follows:
\begin{equation}
\underset{\mathbf{W}_k,\,\mathbf{Z}_k,\, \mathbf{Z}}{min}\sum_{k=1}^K||\mathbf{X}_k\mathbf{W}_k-\mathbf{Z}_k||_F^2 +\beta_k||\mathbf{W}_k||_{2,1}+\zeta_k||\mathbf{Z}_k-\mathbf{Z}||_F^2+\eta||\mathbf{Z}-\mathbf{Y}||_F^2,
\label{MVL}
\end{equation}
where $\mathbf{W}_k\in \mathbb{R}^{d_k\times C}$ is the transformation matrix, $\mathbf{Z}_k\in\mathbb{R}^{N\times C}$ serves as a pseudo-label matrix for $\mathbf{X}_k$, $\mathbf{Z}\in \mathbb{R}^{N\times C}$ is a common matrix for all views, and $||\cdot||_{2,1}$ denotes the $\mathscr{l}_{2,1}$ norm of a matrix, which is defined as the sum of the $\mathscr{l}_{2}$ norms of all column vectors of the matrix and can promote row-sparsity for feature selection~\cite{nie2010efficient}. Note that the first term projects multi-view data to a new space by using transformation matrix $\mathbf{W}_k$. The second term is used as a regularization function. The third and fourth term determines how $\mathbf{Z}_k$ is similar to $\mathbf{Z}$ and how $\mathbf{Z}$ is similar to $\mathbf{Y}$, respectively.

Our aim is to get $\mathbf{W}_k$ from \eqref{MVL}, $k=1,\cdots,K$, so when new testing data $\mathbf{X}^{test} = \{\mathbf{X}_1^{test}, \cdots, \mathbf{X}_K^{test}\}$ comes, we can use $\mathbf{W}_k$ to transform the data matrix $\mathbf{X}_k^{test}$ to a label matrix. The objective function in the testing phase can be written as follows:
\begin{equation}
\underset{\mathbf{Z}^{test}_k,\,\mathbf{Z}^{test}}{min}\sum_{k=1}^K||\mathbf{X}^{test}_k\mathbf{W}_k-\mathbf{Z}^{test}_k||_F^2 +\zeta_k||\mathbf{Z}^{test}_k-\mathbf{Z}^{test}||_F^2, 
\label{MVL_test}
\end{equation}

Finally, $\mathbf{Z}^{test}$ will serve as a label matrix to facilitate the federated multi-view learning.

\subsection{Optimization}
We illustrate how to solve \eqref{MVL} in the training phase and how to solve \eqref{MVL_test} in the testing phase, respectively. Since the objective functions are non-convex and potentially non-smooth, we iteratively update the parameters one by one while fixing the other parameters \cite{feng2020multi}.
The overall algorithm is summarized in Algorithm~\ref{alg: MVL}.


\begin{algorithm}[t]
\SetAlgoLined
\textbf{\emph{Training Phase}}\\
\SetKwInOut{Input}{Input}
\SetKwInOut{Output}{Output}
\Input{\ Dataset $\mathbf{X}=\{\mathbf{X}_1, \cdots,\mathbf{X}_K\}$, and one-hot label matrix $\mathbf{Y}$.}
\Output{\ Transformation matrices $\{\mathbf{W}_k\}$, $k=1,\,\cdots\,,K$.}
Initialize each $\mathbf{W}_k$ randomly; initialize each $\mathbf{Z}_k$ and $\mathbf{Z}$ randomly as $\mathbf{Z}_k^T\mathbf{Z}_k=\mathbf{I}$ and $\mathbf{Z}^T\mathbf{Z}=\mathbf{I}$\;
\While{not converge}{
\For{each $k$ in $[1,K]$}{
\While{not converge}{
Update $\mathbf{A}_k$ according to \eqref{eqUpdateA}\;
Update $\mathbf{W}_k$ according to \eqref{eqUpdateW};
}
Update $\mathbf{Z}_k$ according to \eqref{eqUpdateZk};
}
Update $\mathbf{Z}$ according to \eqref{eqUpdateZ}\;
}
\hrulefill\\
\textbf{\emph{Testing Phase}}\\
\SetKwInOut{Input}{Input}
\SetKwInOut{Output}{Output}
\Input{\ Testing dataset $\mathbf{X}^{test}=\{\mathbf{X}_1^{test},...,\mathbf{X}_K^{test}\}$. Transformation matrices $\{\mathbf{W}_k\}$ $(k=1,...,K)$, which is the output of training phase.}
\Output{\ $\mathbf{Z}^{test}$}
Initialize $\mathbf{Z}_k^{test}$ as $\mathbf{Z}_k^{test}=\mathbf{X}_k^{test}\mathbf{W}_k$\;
\While{not converge}{
Update $\mathbf{Z}^{test}$ according to \eqref{eqUpdateZtest}\;
\For{each $k$ in $[1,K]$}{
Update $\mathbf{Z}_k^{test}$ according to \eqref{eqUpdateZktest}\;
}
}
\caption{Multi-View Learning (MVL)}
\label{alg: MVL}
\end{algorithm}

\subsubsection{Training Phase}\hfill 

In training phase, all parameters are iteratively updated according to the following three steps until \eqref{MVL} converges or reaches a maximum number of iterations.

\textbf{Step 1: update $\mathbf{W}_k$.} When $\mathbf{Z}_k$ and $\mathbf{Z}$ are fixed, the optimization problem of minimizing \eqref{MVL} over $\mathbf{W}_k$ can be written as

\begin{equation}
\underset{\mathbf{W}_k}{min}\sum_{k=1}^K||\mathbf{X}_k\mathbf{W}_k-\mathbf{Z}_k||_F^2 +\beta_k||\mathbf{W}_k||_{2,1}
\label{fix z and zk}
\end{equation}

Following \cite{hou2013joint}, \eqref{fix z and zk} can be written as 

\begin{equation}
\underset{\mathbf{W}_k,\,\mathbf{A}_k}{min}||\mathbf{X}_k\mathbf{W}_k-\mathbf{Z}_k||_F^2 +\beta_k\mathrm{Tr}(\mathbf{W}_k^T\mathbf{A}_k\mathbf{W}_k),
\label{eqW}
\end{equation}
where $\mathbf{A}_k\in\mathbb{R}^{d_k\times d_k}$ is a diagonal matrix with diagonal elements
\begin{equation}
\mathbf{A}_k^{(i,j)} = \left\{
\begin{array}{cc}
1/[2(||\mathbf{W}_k^{(i)}||_2+\epsilon)],&i=j \\ 0, &i\neq j 
\end{array}
\right.\label{eqUpdateA}
\end{equation}

Then, when $\mathbf{A}_k$ is fixed, $\mathbf{W}_k$ can be updated by:
\begin{equation}
    \mathbf{W}_k = (\mathbf{X}_k^T\mathbf{X}_k + \beta_k\mathbf{A}_k)^{-1}\mathbf{X}_k^T\mathbf{Z}_k \label{eqUpdateW}
\end{equation}

So when $\mathbf{Z}_k$ and $\mathbf{Z}$ are fixed, we can repeat the following operation to get $\mathbf{W}_k$ until the value of \eqref{fix z and zk} converges: fix $\mathbf{W}_k$ to update $\mathbf{A}_k$ through \eqref{eqUpdateA} and then fix $\mathbf{A}_k$ to update $\mathbf{W}_k$ through \eqref{eqUpdateW}.

\textbf{Step 2: update $\mathbf{Z}_k$.}
When $\mathbf{Z}$ and $\mathbf{W}_k$ are fixed, \eqref{MVL} becomes:
\begin{equation}
    \underset{\mathbf{Z}_k}{min}\sum_{k=1}^K||\mathbf{X}_k\mathbf{W}_k-\mathbf{Z}_k||_F^2 +\zeta_k||\mathbf{Z}_k-\mathbf{Z}||_F^2  \label{eqZk}
\end{equation}

Then $\mathbf{Z}_k$ can be updated directly by:
\begin{equation}
\mathbf{Z}_k = (\mathbf{X}_k\mathbf{W}_k + \zeta_k\mathbf{Z})/(1+\zeta_k) \label{eqUpdateZk}
\end{equation}

\textbf{Step 3: update $\mathbf{Z}$.}
When $\mathbf{Z}_k$ and $\mathbf{W}_k$ are fixed, \eqref{MVL} becomes:
\begin{equation}
    \underset{\mathbf{Z}}{min}\sum_{k=1}^K \zeta_k||\mathbf{Z}_k-\mathbf{Z}||_F^2+\eta||\mathbf{Z}-\mathbf{Y}||_F^2  \label{eqZ}
\end{equation}

Then $\mathbf{Z}$ can be updated directly by:
\begin{equation}
\mathbf{Z} = (\sum_{k=1}^K\zeta_k\mathbf{Z}_k + \eta\mathbf{Y})/(\sum_{k=1}^K\zeta_k + \eta) \label{eqUpdateZ}
\end{equation}

\begin{figure}[t]
\centering
\includegraphics[width=13.5cm]{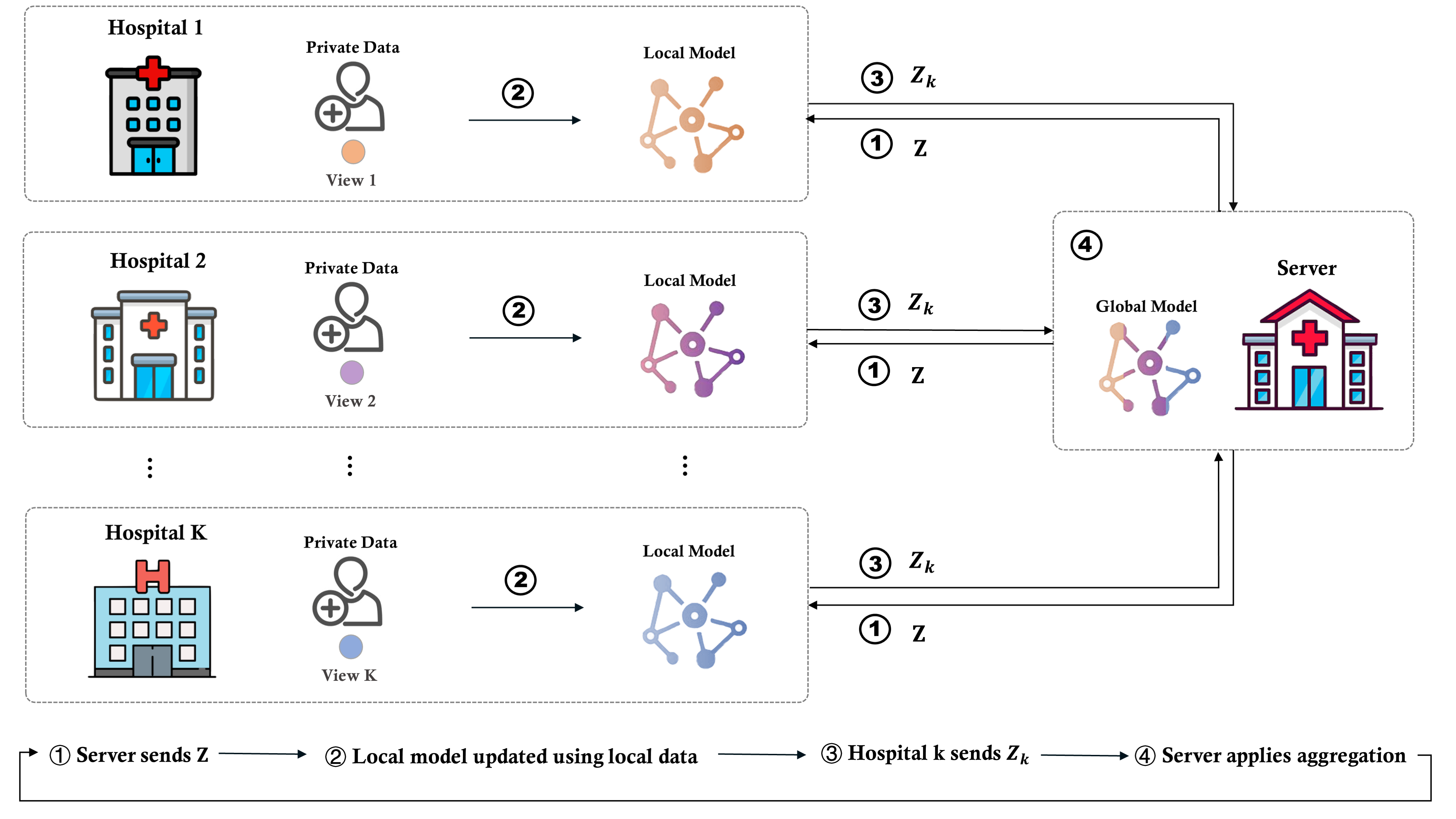}
\caption{Vertical Federated Multi-View Learning (V-FedMV)}
\label{fig:Fed_V}
\end{figure}

\subsubsection{Testing Phase} \hfill

Firstly, initialize $\mathbf{Z}^{test}_k$ as  $\mathbf{Z}^{test}_k = \mathbf{X}^{test}_k\mathbf{W}_k$.
When $\mathbf{Z}^{test}_k$ is fixed, \eqref{MVL_test} can be written as:
\begin{equation}
\underset{\mathbf{Z}^{test}}{min}\sum_{k=1}^K\zeta_k||\mathbf{Z}^{test}_k-\mathbf{Z}^{test}||_F^2 \label{eqZtest}
\end{equation}

Then $\mathbf{Z}^{test}$ can be updated directly by:
\begin{equation}
\mathbf{Z}^{test} = \sum_{k=1}^K\zeta_k\mathbf{Z}^{test}_k/\sum_{k=1}^K\zeta_k \label{eqUpdateZtest}
\end{equation}

When $\mathbf{Z}^{test}$ is fixed, Then $\mathbf{Z}^{test}_k$ can be updated directly by:
\begin{equation}
\mathbf{Z}^{test}_k = (\mathbf{X}^{test}_k\mathbf{W}_k+\zeta_k\mathbf{Z}^{test})/(1+\zeta_k) \label{eqUpdateZktest}
\end{equation}

\section{VERTICAL FEDERATED MULTI-VIEW LEARNING} \label{sec:v_fedmv}

\begin{algorithm}[t]
\SetAlgoLined
\textbf{\emph{Training Phase}}\\
\SetKwInOut{Input}{Input}
\SetKwInOut{Output}{Output}
\Input{\ $\mathbf{X}_k$ that is owned by $L_k$. One-hot label matrix Y that is shared by all $L_k$ and the server, $k=1,\cdots,K$.}
\Output{\ Transformation matrix $\mathbf{W}_k$ that is owned by $L_k$, $k=1,\cdots,K$.}
Each $L_k$ initializes each $\mathbf{W}_k$ randomly\; 
Each $L_k$ initializes each $\mathbf{Z}_k$ randomly as $\mathbf{Z}_k^T\mathbf{Z}_k=\mathbf{I}$\; 
Server initializes $\mathbf{Z}$ as $\mathbf{Z}^T\mathbf{Z}=\mathbf{I}$\;
\While{not converge}{
Server sends $\mathbf{Z}$ to each $L_k$\;
\For{each $L_k$, $k=1,\cdots,K$ \textbf{in parallel}}{
\While{not converge}{
Update $\mathbf{A}_k$ according to \eqref{eqUpdateA} on $L_k$\;
Update $\mathbf{W}_k$ according to \eqref{eqUpdateW}  on $L_k$;
}
Update $\mathbf{Z}_k$ according to \eqref{eqUpdateZk} on $L_k$;
}
Each $L_k$ sends $\zeta_k$ and $\mathbf{Z}_k$ to Server\;
Update $\mathbf{Z}$ according to \eqref{eqUpdateZ} on Server\;
}
\hrulefill\\
\textbf{\emph{Testing Phase}}\\
\SetKwInOut{Input}{Input}
\SetKwInOut{Output}{Output}
\Input{\ Testing dataset $\mathbf{X}_k^{test}$ and transformation matrix $\mathbf{W}_k$ that is owned by $L_k$, $k=1,\cdots,K$.}
\Output{\ $\mathbf{Z}^{test}$}
Each $L_k$ initializes $\mathbf{Z}_k^{test}$ as $\mathbf{Z}_k^{test}=\mathbf{X}_k^{test}\mathbf{W}_k$\;
\While{not converge}{
Each $L_k$ sends $\zeta_k$ and $\mathbf{Z}_k^{test}$ to Server\;
Update $\mathbf{Z}^{test}$ according to \eqref{eqUpdateZtest} on Server\;
\For{each $k$ in $[1,K]$ \textbf{in parallel}}{
Server sends $\mathbf{Z}^{test}$ to each $L_k$\;
Update $\mathbf{Z}_k^{test}$ according to \eqref{eqUpdateZktest} on $L_k$\;
}
}
\caption{Vertical Federated Multi-View Learning (V-FedMV)}
\label{alg:v-fedmv}
\end{algorithm}

In this section, we elaborate on how proper design of the above multi-view learning mechanism can yield guaranteed vertical federated multi-view learning (V-FedMV) gains which consequently not only allows to connect multi-view relations over a global server for remote collaboration, but also to set up an environment across multiple decentralized data to distribute the model-training to the data-owners, without sharing data between institutions.



Fig.~\ref{fig:Fed_V} provides an overview of the V-FedMV framework for vertical accumulation of different views of data for multi-view classification. The overall algorithm is summarized in Algorithm~\ref{alg:v-fedmv}. Briefly, the main procedures of V-FedMV can be described as follows: We first establish a server with multiple distributed clients, then the computations involving the original sensitive data will be executed on clients, while some other computations related to insensitive data of all clients will be executed on the server. For example, \eqref{eqUpdateA}, \eqref{eqUpdateW}, \eqref{eqUpdateZk} and \eqref{eqUpdateZktest} can be computed on each client $L_k$, while \eqref{eqUpdateZ} and \eqref{eqUpdateZtest} can be computed on the server.

\section{HORIZONTAL FEDERATED MULTI-VIEW LEARNING}  \label{sec:h_fedmv}
\begin{figure}[t]
\centering
\includegraphics[width=13.5cm]{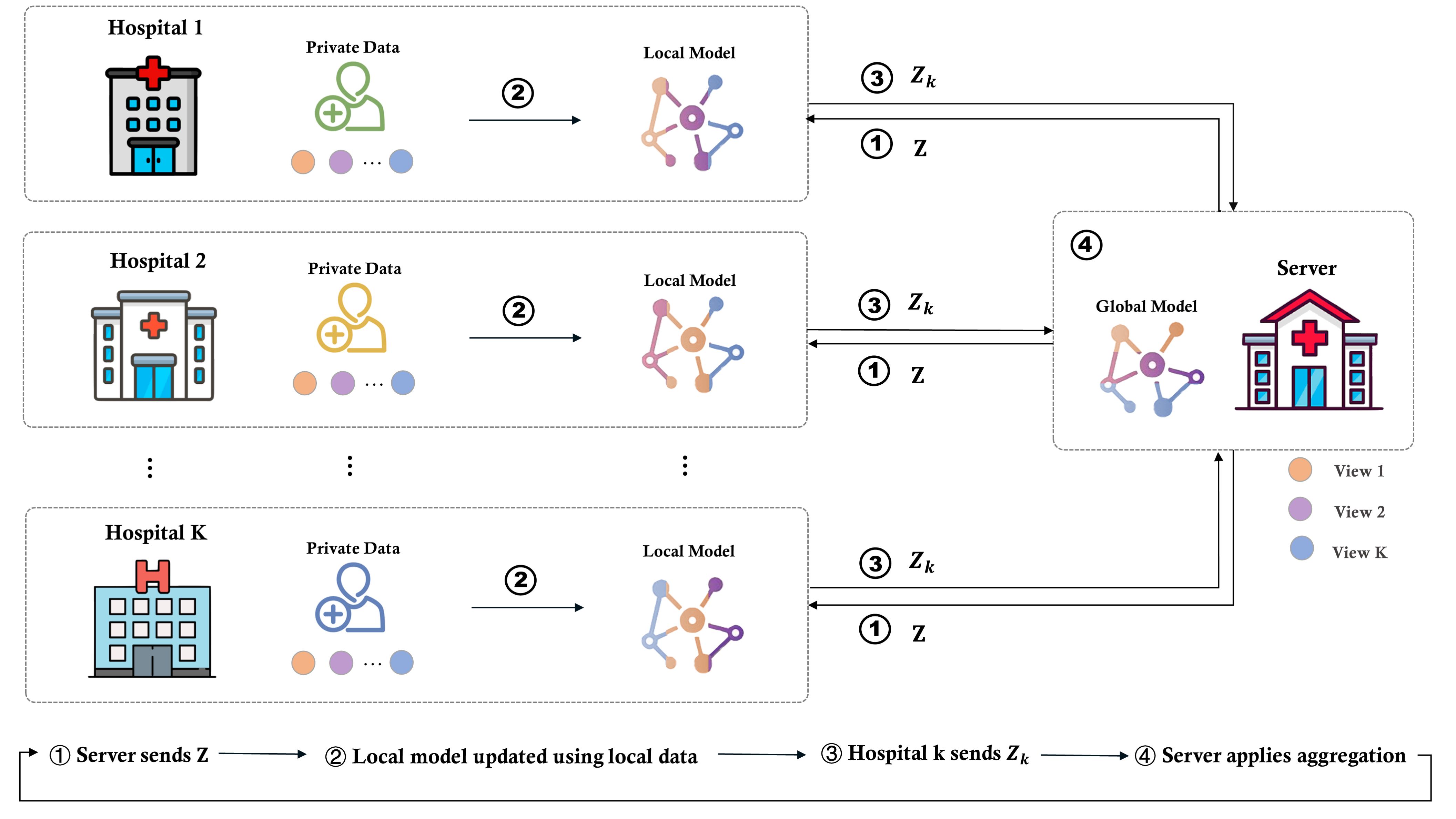}
\caption{Horizontal Federated Multi-View Learning (H-FedMV)}
\label{fig:Fed_H}
\end{figure}

In this section, we illustrate how multi-view learning mechanism can be used for horizontal federated multi-view learning (H-FedMV), which consequently allows to leverage all available multi-view samples from a collection of decentralized local data over a central global server for remote collaboration, but without delivering any raw data to the global server. Fig.~\ref{fig:Fed_H} provides an overview of the H-FedMV framework for horizontal accumulation of different samples of data for multi-view classification. 

Before proceeding, we first redefine the notation, since the setting of H-FedMV is different from V-FedMV. Suppose there are $M$ clients, each of which owns multi-view data but has different patients, denoted by $\{L_l\}_{l = 1}^M$. Let $\mathbf{X}^l_k \in \mathbb{N_l \times d_k}$ denote the $k$-th single-view data on the $l$-th client, where $n_l$ represents the number of samples on $l$-th client. Then the data owned by $L_l$ is defined as $\mathbf{X}^l=\{\mathbf{X}^l_1,\cdots,\mathbf{X}^l_K\}$. The goal of H-FedMV is to use more samples from different institutes or hospitals without data sharing for training models to increase the spectrum of patients incorporated in learning, thereby giving better predictive performance than the individual local averages and preserving data privacy. 


\begin{algorithm}
\SetAlgoLined
\textbf{\emph{Training Phase}}\\
\SetKwInOut{Input}{Input}
\SetKwInOut{Output}{Output}
\Input{\  Dataset $\mathbf{X}^l=\{\mathbf{X}^l_1,\cdots,\mathbf{X}^l_K\}$ and one-hot label matrix $\mathbf{Y}^l$ on $L_l$, $(l=1,2,...,M)$.}
\Output{\ Transformation matrices $\{\mathbf{W}_k\}$, $k=1,...,K$.}
Server initializes $\{\mathbf{W}_k\}$ randomly, $k=1,...,K$ \; 
Each $L_l$ initializes each $\mathbf{Z}_k^{l}$ randomly as $(\mathbf{Z}_k^l)^T\mathbf{Z}_k^l=\mathbf{I}$\; 
Each $L_l$ initializes each $\mathbf{Z}^l$ randomly as $(\mathbf{Z}^l)^T\mathbf{Z}^l=\mathbf{I}$\; 
\For{round in $1,2,...,max\_rounds$}{
Server sends $\{\mathbf{W}_k\},k=1,\cdots,K$ to each $L_l$\;
\For{each $L_l$, $l=1,\cdots,M$ \textbf{in parallel}}{
\While{not converge}{
\For{$k=1,\cdots,K$}{
Update $\mathbf{Z}_k^l$ according to \eqref{eqUpdateZkH} on $L_l$\;
}
Update $\mathbf{Z}^l$ according to \eqref{eqUpdateZH} on $L_l$\;
\For{$k=1,...,K$}{
\While{not converge}{
Update $\mathbf{A}_k^l$ according to \eqref{eqUpdateAH} on $L_l$\;
Update $\mathbf{W}_k^l$ according to \eqref{eqUpdateWH}  on $L_l$;
}
}
}
$L_l$ sends $\{\mathbf{W}_1^l,\mathbf{W}_2^l, \cdots ,\mathbf{W}_K^l\}$ to Server.
}
\For{$k=1,\cdots, K$}{
Server computes $\mathbf{W}_k$ as
$\mathbf{W}_k=\sum_{l=1}^M\frac{N_l}{N}\mathbf{W}_k^l$.
}
}
\hrulefill\\
\textbf{\emph{Testing Phase}}\\
\SetKwInOut{Input}{Input}
\SetKwInOut{Output}{Output}
\Input{\ Testing dataset $\mathbf{X}^{l,test} = \{\mathbf{X}^{l,test}_1, \cdots,\mathbf{X}^{l,test}_K\}$ which is owned by $L_l$, $l=1,2,\cdots,M$. Transformation matrices $\{\mathbf{W}_k\}$, $k=1,\cdots,K$, which is the output of training phase.}
\Output{\ $\mathbf{Z}^{l,test}$}
Each $L_l$ initializes $\mathbf{Z}_k^{l,test}$ as $\mathbf{Z}_k^{l,test} = \mathbf{X}_k^{l,test}\mathbf{W}_k$\;
\For{each $L_l$, $l=1,\cdots,M$ \textbf{in parallel}}{
\While{not converge}{
Update $\mathbf{Z}^{l,test}$ according to \eqref{eqUpdateZtestH} on $L_l$\;
\For{each $k$ in $[1,K]$}{
Update $\mathbf{Z}_k^{l,test}$ according to \eqref{eqUpdateZktestH} on $L_l$\;
}
}
}
\caption{Horizontal Federated Multi-View Learning (H-FedMV)}
\label{alg:h-fedmv}
\end{algorithm}

After introducing the notation, it is easy to extend the multi-view learning mechanism to the horizontal federated learning case, which we just need to add the superscript $l$ to each equation to indicate that the data is owned by the $l$-th local organization. Accordingly, the Eqs.~ \eqref{eqUpdateA}, \eqref{eqUpdateW}, \eqref{eqUpdateZk}, \eqref{eqUpdateZ}, \eqref{eqUpdateZtest}, \eqref{eqUpdateZktest} can be set up as:

The overall algorithm of H-FedMV is summarized in Algorithm~\ref{alg:h-fedmv}. Briefly, the main procedures can be described as follows: firstly, all clients apply the algorithm of the multi-view learning on their own data and devices, then each of them can get $\mathbf{W}_k^l$, which are sent to the global server to compute the weighted average value. After the computation on server, it sends the output weighted average transformation matrix $\mathbf{W}_k$  back to the local client $L_l$ for distributed model training. This process will be repeated for several times until convergence is reached or some stopping criteria are met. At last, the $\mathbf{W}_k$ of all $K$ views are obtained on the server and then server will send them to clients, which can be used to predict testing data of each client.

\begin{equation}
\mathbf{A}_k^{l (i,j)} = \left\{
\begin{array}{cc}
\frac{1}{2(||\mathbf{W}_k^{l (i)}||_2+\epsilon)}, &i=j \\ 0, &i\neq j 
\end{array}
\right.\label{eqUpdateAH}
\end{equation}

\begin{equation}
    \mathbf{W}_k^l = [(\mathbf{X}_k^l)^{ T}\mathbf{X}_k^l + \beta_k^l\mathbf{A}_k^{l}]^{-1}(\mathbf{X}_k^l)^ {T}\mathbf{Z}_k^l \label{eqUpdateWH}
\end{equation}

\begin{equation}
\mathbf{Z}_k^l = (\mathbf{X}_k^l\mathbf{W}_k^l + \zeta_k^l\mathbf{Z}^l)/(1+\zeta_k^l) \label{eqUpdateZkH}
\end{equation}

\begin{equation}
\mathbf{Z}^l = (\sum_{k=1}^K\zeta_k^l\mathbf{Z}_k^l + \eta^l\mathbf{Y}^l)/(\sum_{k=1}^K\zeta_k^l + \eta^l) \label{eqUpdateZH}
\end{equation}

\begin{equation}
\mathbf{Z}^{l,test} = \sum_{k=1}^K\zeta_k^l\mathbf{Z}^{l,test}_k/\sum_{k=1}^K\zeta_k^l \label{eqUpdateZtestH}
\end{equation}

\begin{equation}
\mathbf{Z}^{l,test}_k = (\mathbf{X}^{l,test}_k\mathbf{W}_k+\zeta_k^l\mathbf{Z}^{l,test})/(1+\zeta_k^l) \label{eqUpdateZktestH}
\end{equation}

\section{FEDERATED MULTI-VIEW SEQUENTIAL LEARNING} \label{sec:sequential}

\begin{figure}[t]
\centering
\includegraphics[width=13.5cm]{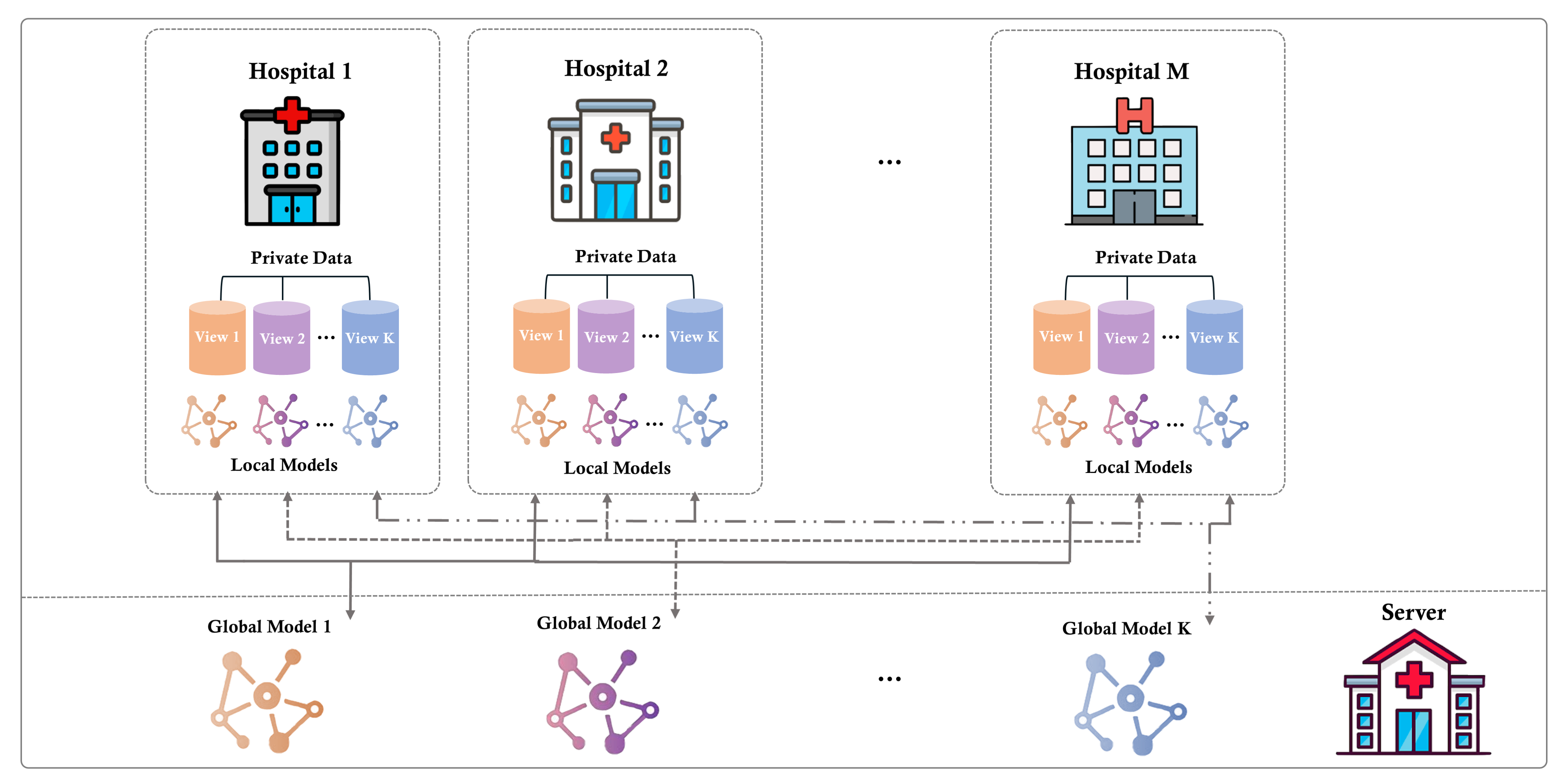}
\caption{Federated Multi-view Sequential Learning}
\label{fig:fed_seq_prep}
\end{figure}

In medical practice, sequential or longitudinal data is very common, such as clinical research and epidemiological studies, 
However, V-FedMV and H-FedMV cannot be directly applied to sequential data for considering the temporal information of the multi-view data. We further propose a federated multi-view sequential learning method to adapt our model to deal with multi-view sequential data in a federated environment.

In the vertical setting, each client contains a single-view data, and can be independently executed (such as using GRU) on their own data to get the feature embedding matrix from sequential data and then apply the proposed V-FedMV method, which will not cause expose the raw data in the whole process. However, in the horizontal setting, each client owns multi-view data partially and they cannot process their own data independently, because the trained GRU models for each view are inconsistent locally.  
Hence, how to do federated multi-view learning in the horizontal setting when the raw data is sequential is a problem. Here we develop a sequential version of H-FedMV to illustrate how multi-view sequential data can be arranged by using federated learning, named S-FedMV.


\begin{algorithm}[t]
\SetAlgoLined
\SetKwInOut{Input}{Input}
\SetKwInOut{Output}{Output}
\SetKwFunction{FLT}{\textbf{LocalTraining}}
\SetKwProg{Fn}{Function}{:}{}
\Input{\ Sequential data $\{D^l_k\}$, $l=1,\cdots,M$, $k=1,\cdots,K$}
\Output{\ $\{\mathbf{X}^l_k\}$, $l=1,\cdots,M$, $k=1,\cdots,K$}
\For{$k$-th view $(k=1,\cdots,K)$}{
Server initializes $\mathbf{w}_0^k$\;
\For{each round $t$ = $0, 1, \cdots, max\_rounds$}{
Server send $\mathbf{w}_t^k$ to each client $L_l$ $(l=1, \cdots, M)$\;
\For{each client $L_l$ $(l=1,\cdots,M)$ \textbf{in parallel}}{
$L_l$ computes: $\mathbf{w}_{t+1}^{k,l} \leftarrow$ \FLT{$l$, $k$, $\mathbf{w}_t^k$}\;
$L_l$ sends $\mathbf{w}_{t+1}^{k,l}$ to Server\;
}
Server computes: $\mathbf{w}_{t+1}^k \leftarrow \sum_{l=1}^M\frac{N_l}{N}\mathbf{w}_{t+1}^{k,l}$\; 
}
Server send the last $\mathbf{w}^k$ to each $L_l$\;
$L_l$ uses the last $\mathbf{w}^k$ to get the $k$-th feature matrix $\mathbf{X}_k^l$.
}

\Fn{\FLT{$l$,\,$k$,\,$\mathbf{w}$}}{
$\mathcal{B} \leftarrow$ (split $D^l_k$ into batches of size B)\;
\For{each local epoch $i$ from 1 to $MaxIter$}{
\For{batch b $\in$ $\mathcal{B}$}{
$\mathbf{w}$ $\leftarrow$ $\mathbf{w}$ - $\eta\nabla \mathscr{l}(\mathbf{w};b)$
}
}
\KwRet $\mathbf{w}$
}
\caption{Federated Multi-View Sequential Learning}
\label{FedPreprocessing}
\end{algorithm}


Fig.~\ref{fig:fed_seq_prep} provides an overview of the S-FedMV framework for horizontal accumulation of different samples of multi-view sequential data for feature extraction. Briefly, the idea is based on the Federated Averaging (FedAvg), following a server-client setup with two repeated stages: (i) the clients train their models locally on their data, and (ii) the server collects and aggregates the models to obtain a global model by weighted averaging. For each client, we assign the aggregation weight as the ratio of data samples on each client to the total number of samples, namely $N_l/N$. It is appropriate to process the data of each view separately in the case of
distributed horizontal multi-view data, because there exist inconsistency among the models of different views. FedAvg is flexible to the model and the optimizer used for training, here we use the bidirectional GRU as the model and RMSProp as the optimizer in the experiments. For simplicity, in every round we choose all clients to participate in.

The overall algorithm is summarized in Algorithm~\ref{FedPreprocessing}, and the notations used in the proposed model are as below: As before, we use $\{L_l\}$, $(l=1,...,M)$ to denote $M$ clients. Let $\{D^l_k\}$ denote the sequential data on $l$-th client in the $k$-th view, $\mathbf{w}_t^k$ and $\mathbf{w}_t^{k,l}$ denote the parameters of $k$-th view model on server and on the client in the $t$ round, and $\mathscr{l}(\mathbf{w};b)$ denote the loss of the model with parameters $\mathbf{w}$ and $b$, where $b$ is the bias.

\section{Discussion}\label{sec:privacy}
In V-FedMV, both $\mathbf{W}_k$ and $\mathbf{Z}_k$ are updated on local devices while $\mathbf{Z}$ is updated on the server. $\zeta_k$ and $\mathbf{Z}_k$ are sent to the server from local devices when the server updates $\mathbf{Z}$, which would not leak the raw data because $\mathbf{X}_k$ is preserved at local devices throughout the process. Besides, in some cases, the server may not be set up by these clients, but by another specialized organization that would not provide the raw data but only do the computation, and in this case, although $\mathbf{Y}$ is also needed when server updates $\mathbf{Z}$, it won't cause leakage because $\mathbf{Y}$ is only a one-hot label matrix which would not be useful when don't know extra information.

In H-FedMV and S-FedMV, each client sends the parameters of its own model to the server in each round, and the server averages it and returns the result. Each client then continues to update locally with the parameters returned by the server. In the whole process, the server only contacts the parameters, not the raw data, and the local devices can only contact the updated parameters from the server, not the raw data from other clients, so this ensures privacy.

\section{Experimental Evaluation} \label{sec:exp}

In this section, 
we evaluate the performance of the proposed V-FedMV, H-FedMV, and S-FedMV on a new multi-view sequential keystroke data which is collected from the BiAffect\footnote{https://www.biaffect.com \label{biaffect}} study.

\subsection{Data Description}
BiAffect\textsuperscript{\ref{biaffect}}, the first study on mood and cognition using mobile typing kinematics, provides the multi-view sequential data for our experiments. BiAffect invited 40 participants to use the customized smartphones in their daily life. These phones are equipped with a custom keyboard that would collect the data of keypress duration, typing behaviors, accelerometer value, and others. In the experiment, we use three types of metadata: alphanumeric characters, special characters, and accelerometer values, which can be seen as three views:
\textit{Alphanumeric characters} include the duration time of keypress, time consumed since the last key was pressed, and the distance between the current key and the previous key along two axes;
\textit{Special characters} are auto-correct, backspace, space, suggestion, switching-keyboard, and other special characters;
\textit{Accelerometer values} are collected by the sensor of the smartphone.
The device records accelerometer values every 60ms, regardless of typing speed.

The diagnosis of patients with bipolar disorder was obtained by weekly assessment of participants through Hamilton Depression Rating Scale (HDRS) \cite{williams1988structured} and Young Mania Rating Scale (YMRS) \cite{young1978rating}, which is a very reliable assessment standard for bipolar disorder. There are 7 participants with bipolar I disorder, 5 participants with bipolar II disorder, and 8 participants with no diagnosis per DSM-IV TR criteria~\cite{kessler2005lifetime}.

\subsection{Experimental Setup}
\paragraph{Data Preprocessing} In the experiment, we investigate a session-based mood prediction problem same as prior works in the BiAffect project~\cite{cao2017deepmood}, which utilizes all three views to predict a user's mood score. However, our goal is to train a binary mood classification. To get the binary labels, after consulting with the professional experts, we label the sessions with the HDRS score between 0 and 7 (inclusive) as negative samples and those with the HDRS score higher than score 7 as positive samples.

\begin{figure}[h]
	\centering
	\subfloat[Visualization of labeling with t-SNE for three views after data preprocessing]{\includegraphics[width=13.5cm]{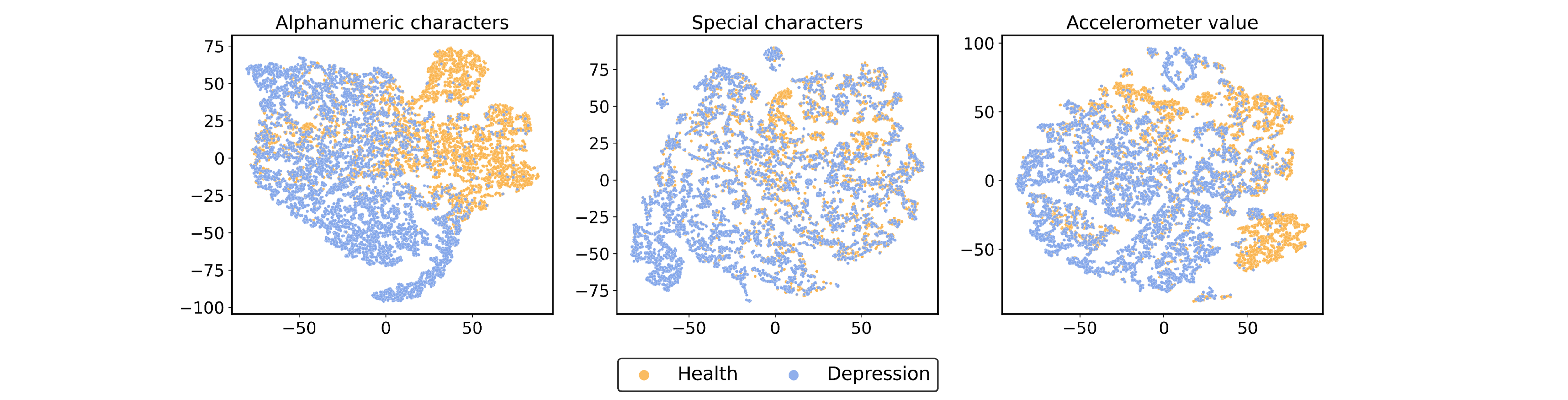}\label{fig:data_after_GRU}}\\
	\subfloat[Visualization of labeling with t-SNE for three views after V-FedMV]{\includegraphics[width=13.5cm]{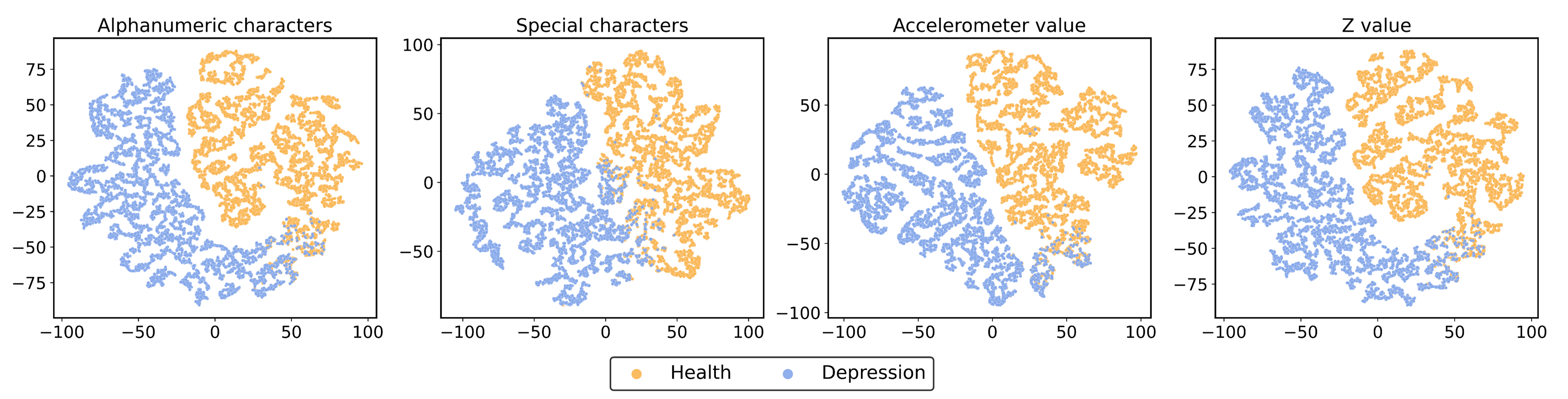}\label{fig:data_after_V-FedMV}}\\
	\subfloat[Visualization of labeling with t-SNE for three views after H-FedMV]{\includegraphics[width=13.5cm]{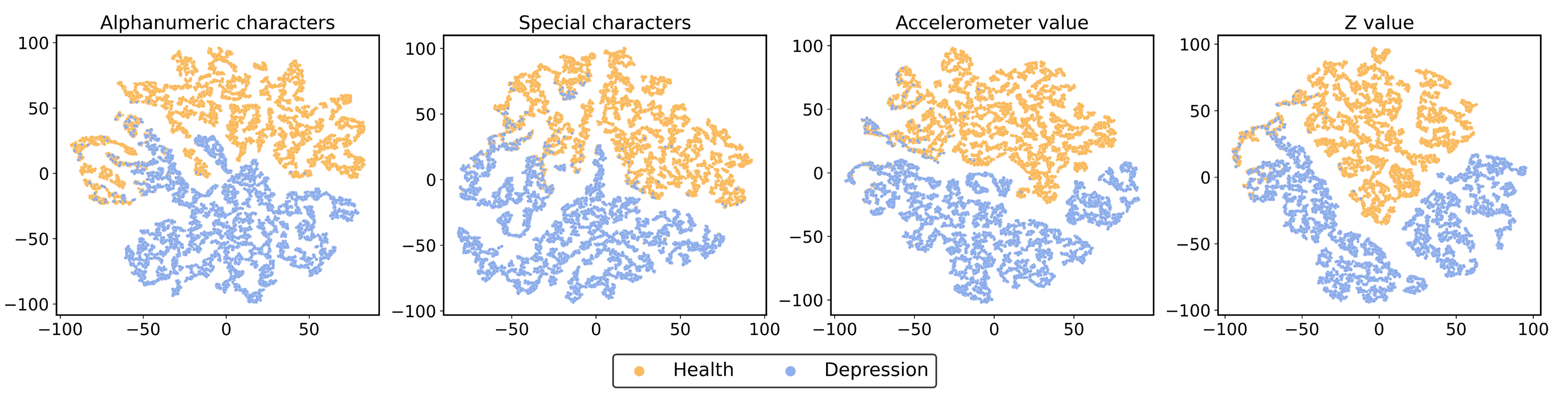}\label{fig:data_after_H-FedMV}} \\
	\caption{Summary of visualization in various settings.} \label{fig:visualization}
\end{figure}

However, the original input data is not naturally fitted our proposed V-FedMV and H-FedMV, because the original keystroke data is time-sequential format, but these methods are designed for feature matrices. To address the data format problem, we applied a GRU~\cite{cho2014learning} which is a simplified version of Long Short-Term Memory (LSTM)~\cite{hochreiter1997long} to each view respectively to preprocess the data and extract feature embedding matrix as the input of V-FedMV and H-FedMV methods.

In this project, we use Keras with Tensorflow as the backend to implement the code. We use RMSProp~\cite{tieleman2012lecture} as the optimizer for GRU training. We retain sessions that any view contains the number of keypress between 10 and 100, and then we have 14,971 total samples. Besides, we set the batch size as 128, the epoch as 500, the learning rate from $\{0.001,0.005\}$, and the dropout from $\{0.1,0.3\}$. We use the validation dataset to select the optimal parameters and get the feature matrices which are the input of our V-FedMV and H-FedMV frameworks. Furthermore, the visualization of three views after data preprocessing is shown in Fig.~\ref{fig:data_after_GRU}, demonstrating the original space of the input data of V-FedMV and H-FedMV. The visualizations of each view and $\mathbf{Z}$ from V-FedMV and H-FedMV are shown in Fig.~\ref{fig:data_after_V-FedMV} and Fig.~\ref{fig:data_after_H-FedMV}, respectively. It is not hard to see after V-FedMV and H-FedMV, the data can be well classified.


In addition, we conduct various experiments to evaluate the performance of S-FedMV. The max communication round is 150 for alphanumeric and special views, 170 for accel view. We set the local epoch from $\{10,15\}$, the dropout from $\{0.1,0.3\}$. and the batch size as 128. The same as previous approaches, we select the optimal parameters through validation and finally get the best feature matrices.

\paragraph{Hyperparameter Setting} 
After data preprocessing, we carry out experiments on V-FedMV and H-FedMV. In V-FedMV, we set up three clients, split the processed feature matrices according to the type of view, and make each client own one of three views. In H-FedMV, we set four clients, split the processed feature matrices, and make each client own the same number of samples with all three views. One more thing is that we also equally assigned positive and negative samples to each client for H-FedMV.

For more details of hyperparameter setting, V-FedMV uses $\beta_k$ as 4, $\zeta_k$ and $\eta$ from $2^0$ to $2^5$. In H-FedMV, we set the max communication rounds as 20, and the value of $\beta_k^l$ as 4, and $\zeta_k^l$, $\eta^l$ across different clients to be the same from $2^0$ to $2^5$. For both V-FedMV and H-FedMV, the optimal parameters are also selected by validation. We repeat all the experiments ten times with four metrics, i.e., accuracy, precision, recall, and f1 score. In addition, we also calculate the average and standard deviation.




\subsection{Vertical Federated Multi-View Learning}

\begin{table}[t]
\centering
\begin{tabular}{l| c| c| c| c} 
\hline
Metric & Accuracy(\%) & Precision(\%) & Recall(\%) & F1(\%)\\ [0.5ex]
\hline
V-FedMV & \textbf{88.76 $\pm$ 0.74}   &  \textbf{90.89 $\pm$ 0.83}  & 87.09 $\pm$ 1.89   & \textbf{88.93 $\pm$ 0.84}  \\ 
\hline
Pairwise FL w/o Special & 88.30 $\pm$ 0.64 & 89.54 $\pm$ 0.68 & \textbf{87.72 $\pm$ 1.47} & 88.61 $\pm$ 0.70 \\
\hline
Pairwise FL w/o Accel &  84.92 $\pm$ 0.50  &  89.00 $\pm$ 1.34  & 81.01 $\pm$ 2.00   &  84.79$\pm$ 0.67 \\
\hline
Pairwise FL w/o Alphanum & 78.71 $\pm$ 0.74   & 79.63 $\pm$ 1.59  &  79.42 $\pm$ 3.34  & 79.45$\pm$ 1.11  \\  
\hline
 Alphanum w/o FL & 84.53 $\pm$ 0.68  &  87.58 $\pm$ 0.88  &  81.81 $\pm$ 1.40  & 84.59 $\pm$ 0.75  \\
\hline
 Accel w/o FL &  76.94 $\pm$ 0.91  & 75.51 $\pm$ 1.03  &  82.32 $\pm$ 2.73  & 78.73 $\pm$ 1.13 \\ 
\hline
Special w/o FL & 62.30 $\pm$ 0.64 & 70.19 $\pm$ 1.81 & 47.83 $\pm$ 3.09 & 56.79 $\pm$ 1.87 \\ 
\hline
\end{tabular}
\caption{Prediction performance of compared methods in V-FedMV experiment.}
\label{table:V-FedMV}
\vspace{-10pt}
\end{table}

\paragraph{Baselines} In this part, we evaluate V-FedMV with six other baselines. All approaches are summarized as follows:
\begin{itemize}
\item \textbf{V-FedMV}: It is our proposed federated multi-view learning approach in the vertical setting, described in Algorithm~\ref{alg:v-fedmv}. Note that, all three views are used in the experiment of V-FedMV.
\item \textbf{Pairwise FL}: These approaches are similar to V-FedMV that also applies the federated multi-view learning algorithm in vertical setting. The difference is that only two views are considered. In this case, we have three experiments with two views, including Pairwise FL w/o Special, Pairwise FL w/o Accel, Pairwise FL w/o Alphanum.
\item \textbf{Single View w/o FL} \cite{zhao2010efficient}: In this approach, we optimize the following objective function on each view:
\begin{equation}
\underset{\mathbf{W}_k}{min}||\mathbf{X}_k\mathbf{W}_k-\mathbf{Y}||_F^2 +\beta_k||\mathbf{W}_k||_{2,1} \label{VerticalSingle}
\end{equation}
where $\mathbf{Y}$ in \eqref{VerticalSingle} refers to the one-hot label matrix. Similar to pairwise FL, we also have three single view w/o FL experiments, such as Alphanum w/o FL, Accel w/o FL, Special w/o FL.
\end{itemize}


\paragraph{Performance}
Experimental results of V-FedMV and its baselines are shown in Table \ref{table:V-FedMV}. It is not hard to see that V-FedMV shows the best performance with 88.76\% accuracy and 88.93\% f1 score. All methods have a low standard deviation, which means all approaches are very stable in the repeated experiments. Meanwhile, we also find the special view makes the most negligible contribution for this task based on the experimental results of Pairwise FL w/o Special and Special w/o FL. Especially, the Pairwise FL w/o Special approach can perform almost the same as V-FedMV. One of the main reasons is that the special view provides less information than the other two views. Besides, we can see that the alphanum view achieves the best accuracy (84.53\%) and f1 score (84.59\%) among all three views, and the accel view is between the others. While comparing all approaches together, we can easily conclude that the model's performance increases with more views, demonstrating that the proposed method is effective for multi-view data. Finally, the proposed approach V-FedMV improves 4.23\% accuracy, 3.31\% precision, 5.28\% recall, and 4.34 \% f1 score than Alphanum w/o FL that is the best single view learning without FL.



\subsection{Horizontal Federated Multi-View Learning} \label{compared_method_h}

\begin{table}[t]
\centering
\begin{tabular}{l| c| c| c| c} 
\hline
Metric & Accuracy(\%) & Precision(\%) & Recall(\%) & F1(\%)\\ [0.5ex]
\hline
H-FedMV & \textbf{88.98 $\pm$ 0.52} & \textbf{91.01 $\pm$ 0.80} & 87.42 $\pm$ 1.33 & \textbf{89.17 $\pm$ 0.57} \\ 
\hline
MV w/o FL & 87.92 $\pm$ 0.51 & 90.21 $\pm$ 1.53 & 86.15 $\pm$ 2.58 & 88.09 $\pm$ 0.72 \\
\hline
Pairwise FL w/o Special & 88.28 $\pm$ 0.65 & 89.67 $\pm$ 0.69 & \textbf{87.52 $\pm$ 1.54} & 88.57 $\pm$ 0.72 \\
\hline
Pairwise FL w/o Accel & 84.72 $\pm$ 1.01 & 89.06 $\pm$ 1.10 & 80.47 $\pm$ 2.40 & 84.52 $\pm$ 1.23 \\
\hline
Pairwise FL w/o Alphanum & 78.77 $\pm$ 0.72 & 79.70 $\pm$ 1.60 & 79.45 $\pm$ 3.17 & 79.51$\pm$ 0.99 \\  
\hline
Single View FL - Alphanum & 84.57 $\pm$ 0.65 & 87.53 $\pm$ 0.79 & 81.97 $\pm$ 1.44 & 84.65 $\pm$ 0.73 \\
\hline
Single View FL - Accel & 76.96 $\pm$ 0.91 &  75.51 $\pm$ 1.07 & 82.39 $\pm$ 2.60 & 78.77 $\pm$ 1.09 \\ 
\hline
Single View FL - Special & 62.26 $\pm$ 0.58 & 70.20 $\pm$ 1.77 & 47.65 $\pm$ 3.02 & 56.67 $\pm$ 1.81 \\ 
\hline
\end{tabular}
\hfill
\caption{Prediction performance of compared methods in H-FedMV experiment.}
\label{table:H-FedMV}
\vspace{-10pt}
\end{table}

\paragraph{Baselines} Here, we evaluate H-FedMV with seven other baselines. All approaches are summarized as follows:
\begin{itemize}
\item \textbf{H-FedMV}: It is our proposed federated multi-view learning approach in the horizontal setting, described in Algorithm~\ref{alg:h-fedmv}. H-FedMV also uses all three views like V-FedMV.
\item \textbf{MV w/o FL}: In this approach, each client applies the multi-view learning method on its local device. All client don't share the parameters to the server for aggregation. Same as H-FedMV, MV w/o FL also uses all three views.
\item \textbf{Pairwise FL}: These approaches also apply the proposed multi-view federated learning approach in the horizontal setting, but with only two views. Thus, we have three experiments with two views, including Pairwise FL w/o Special, Pairwise FL w/o Accel, Pairwise FL w/o Alphanum.
\item \textbf{Single View FL}: In this approach, the data on clients are represented by a single view. All clients optimize Eq.~\eqref{VerticalSingle} on their own devices firstly. Then each client sends its $\mathbf{W}_k$ to the server for aggregation. After the aggregation is done, the global $\mathbf{W}_k$ will be back to upload the local model for all clients. This process will continue until the maximum number of communication rounds is reached. Similar to pairwise FL, we also have three single view FL experiments: Single View FL - Alphanum, Single View FL - Accel, Single View FL - Special.
\end{itemize}

\paragraph{Performance}
Table \ref{table:H-FedMV} shows the experimental results of H-FedMV and its baselines. It can be observed that H-FedMV outperforms the compared baselines with 88.98\% accuracy and 89.17\% f1 score. MV w/o FL trains each model locally and the average performance of all local models is worse than H-FedMV. One of the main reason is that federated learning enables H-FedMV to get more training sample information than MV w/o FL, resulting in a better performance. In addition, the significance of each view in model training is similar to the experimental results of H-FedMV and its baselines in last section, i.e., alphanum view > accel view > special view. Eventually, the proposed H-FedMV improves 1.06\% accuracy, 0.8\% precision, 1.27\% recall, and 1.08\% f1 score compared to MV w/o FL.



\subsection{Federated Multi-View Sequential Learning}
\paragraph{Baselines} We further evaluate the S-FedMV approach with three other baselines. 
\begin{itemize}
\item \textbf{S-FedMV}: It is our proposed federated learning framework for multi-view sequential data, where each client interacts with the global model with H-FedMV.
\item \textbf{Local Sequential H-FedMV}: Each client runs feature representation learning locally with H-FedMV, which indicates no interactions of feature learning between any two clients.
\item \textbf{Local Sequential LocalMV}: There is no federated learning involved in this approach. 
Each client trains the multi-view sequential model by only using the local data.
\item \textbf{Centralized Sequential H-FedMV}: Centralized sequential approach has the access to all training data across all clients for feature leaning with H-FedMV. In general, it should achieve the best performance due to the best feature representation in federated learning.

\end{itemize}

\begin{table}[t]
\begin{tabular}{l|c|c|c|c} 
\hline
Metric & Accuracy(\%) & Precision(\%) & Recall(\%) & F1(\%)\\ [0.5ex]
\hline
S-FedMV& \textbf{88.90 $\pm$ 0.53}   &  89.93 $\pm$ 0.82  &  \textbf{88.54 $\pm$ 1.16}  & \textbf{89.22 $\pm$ 0.55}  \\ 
\hline 
Local Sequential H-FedMV & 68.31 $\pm$ 5.58  &  74.77 $\pm$ 5.39  & 58.10 $\pm$ 9.06   & 65.24 $\pm$ 7.82\\ 
\hline
Local Sequential LocalMV &82.27 $\pm$ 0.91 &84.87 $\pm$ 1.50 & 80.34 $\pm$ 0.44 &82.44 $\pm$ 0.77\\
\hline
Centralized Sequential H-FedMV & 88.98 $\pm$ 0.52 & 91.01 $\pm$ 0.80 & 87.42 $\pm$ 1.33 & 89.17 $\pm$ 0.57 \\
\hline
\end{tabular}
\caption{Prediction performance of compared methods in S-FedMV}
\label{table:Seq}
\vspace{-10pt}
\end{table}


\paragraph{Performance}
Experimental results of different multi-view federated learning methods for sequential data are shown in Table \ref{table:Seq}. First, we can see that S-FedMV achieves outstanding performance with 88.90\% accuracy and 89.22\% f1 score. It demonstrates that S-FedMV is the best approach for distributed multi-view sequential data since its performance is close to centralized sequential H-FedMV. Note that centralized sequential H-FedMV is the up-bound performance of all approaches since the feature representation learning can access all training data. Then, we find the results of local sequential H-FedMV can confirm our statement that it is infeasible for each client to preprocess its data independently. It is because the trained models are inconsistent between different clients, and they obtained a bad representation learning of H-FedMV. We can also see that the proposed approach S-FedMV is better than local sequential LocalMV, which trains the local model on local sequential data only.

\subsection{Hyperparameter Analysis} 

\begin{figure}[t]
	\centering
	\subfloat[Accuracy]{\includegraphics[width=13.7cm]{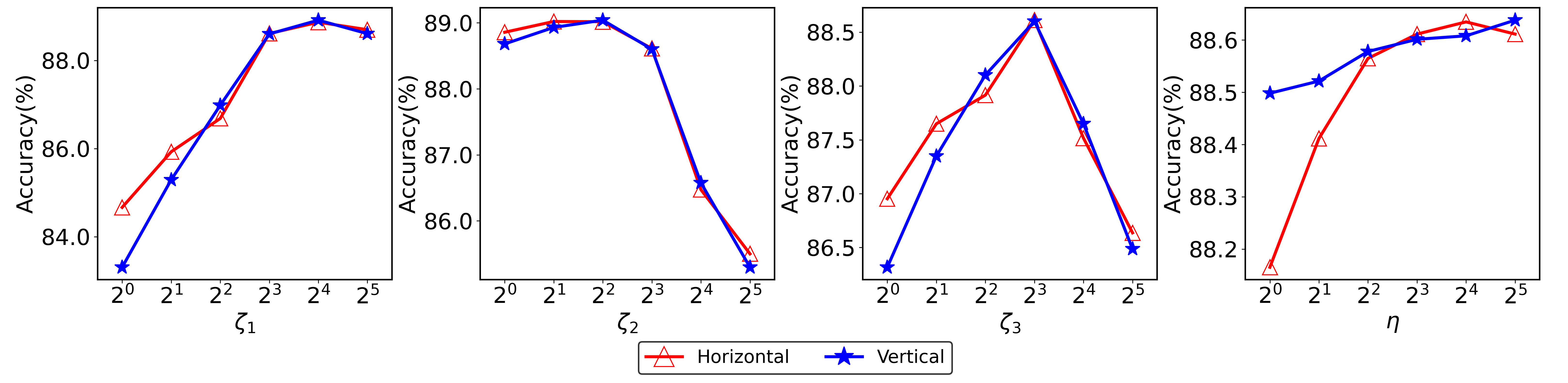}\label{fig:PA_Acc}}\\
	\subfloat[Precision]{\includegraphics[width=13.7cm]{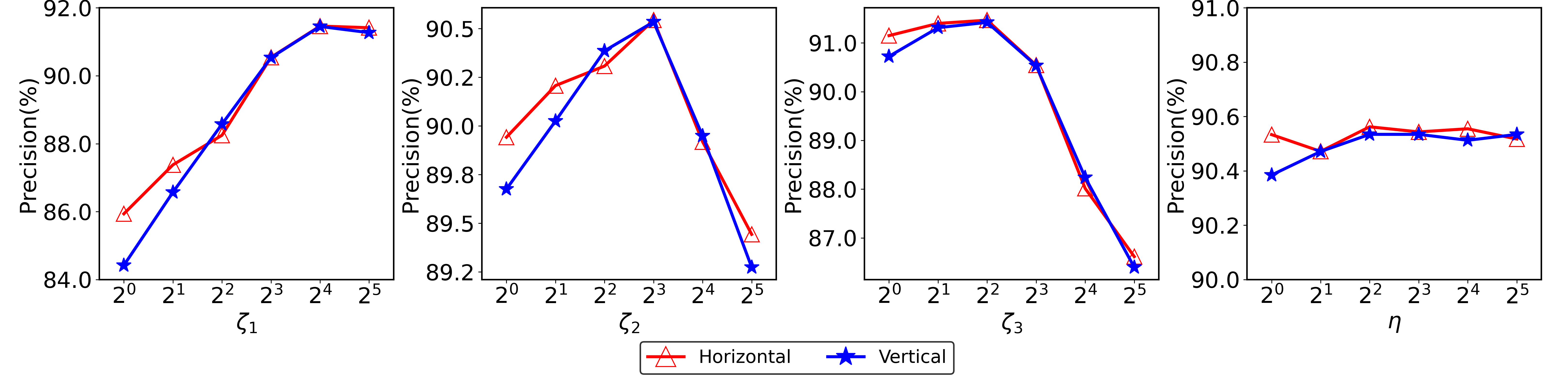}\label{fig:PA_Pre}}\\
	\subfloat[Recall]{\includegraphics[width=13.7cm]{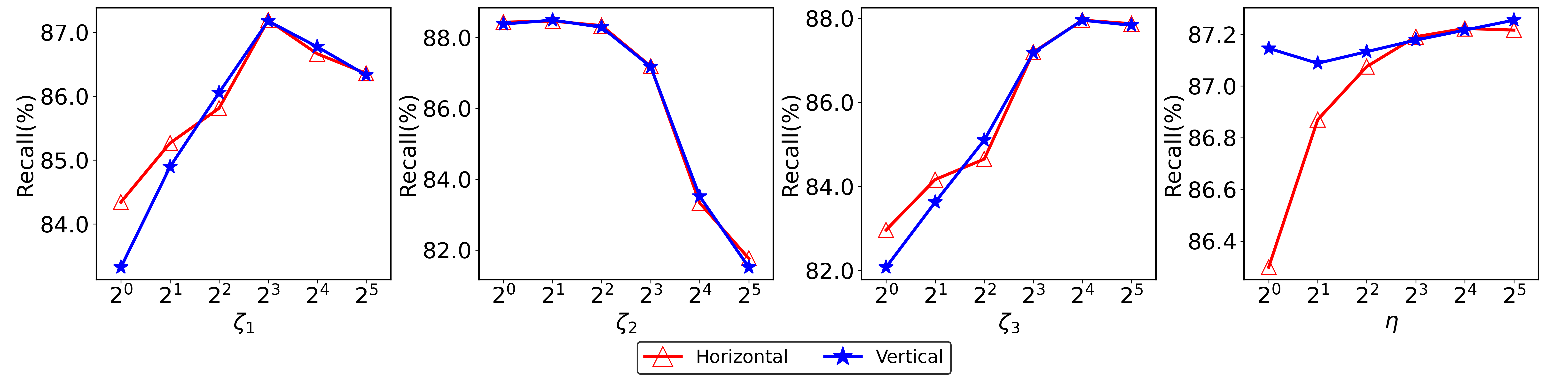}\label{fig:PA_Recall}}\\
	\subfloat[F1]{\includegraphics[width=13.7cm]{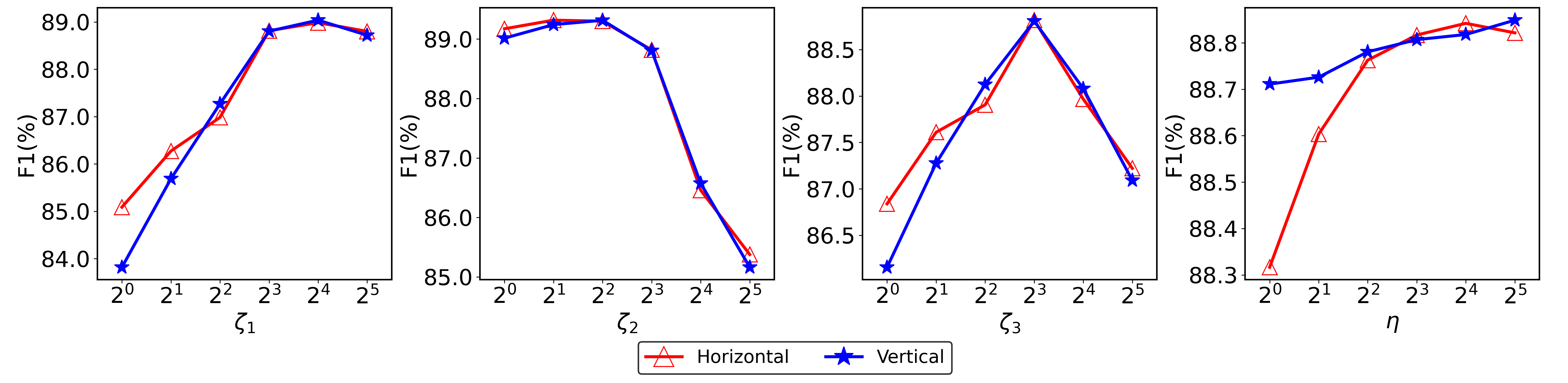}\label{fig:PA_F1}}\\
	\caption{Hyperparameters analysis with V-FedMV and H-FedMV} \label{fig:parameter_analysis}
	\vspace{-10pt}
\end{figure}

We use the hyperparameter $\zeta_1,\zeta_2,\zeta_3$ to correspond with alphanumeric view, special view and accel view respectively, which controls the balance between three views. For instance, a large $\zeta_k$ indicates a higher impact of the $k$-th view of model training. In this experimental setting, we test each hyperparameter $\in \{2^0,2^1,2^2,2^3,2^4,2^5\}$. Note that while testing one hyperparameter, we fix the others as $2^3$. We show the evaluation results with four metrics, i.e., accuracy, precision, recall, f1 scores shown in Fig. \ref{fig:parameter_analysis}. 
We find that the hyperparameters affect the accuracy and f1 score consistently, which shows a similar trend line for the same hyperparameter evaluation.
In summary, while $\zeta_1=2^4,\zeta_2=2^2$ and $\zeta_3=2^3$, the model can achieve the best performance on accuracy and f1 score for both H-FedMV and V-FedMV approaches.
While $\eta=2^4$, H-FedMV is better than V-FedMV, but V-FedMV is better than H-FedMV once $\eta=2^5$.
Moreover, we also find the performance of the trained model is most sensitive to $\zeta_1$ and $\zeta_2$ and less sensitive to $\eta$.








\section{RELATED WORK} \label{sec:related}
 In this section, we review the related work, which can be placed into three main categories: multi-view learning, federated learning, and federated multi-view learning.

\textbf{Multi-View Learning:} Sun \emph{et al.}~\cite{sun2013survey} provided a survey for multi-view machine learning and pointed out that multi-view learning is related to the machine learning problem with the data represented by multiple distinct feature sets. Xu \emph{et al.} analyzed different multi-view algorithms and indicated that it was consensus and complementary principles that ensure their promising performance~\cite{xu2013survey}. The aim of the consensus principle is to minimize the disagreement on multiple distinct views. A connection between the consensus of two hypotheses on two views respectively and their error rates was given by Dasgupta \emph{et al.}~\cite{dasgupta2002pac}. The complementary principle means that multiple views can be complements for each other and can be exploited comprehensively to produce better learning performance for the reason that each view may contain some specific information that other views do not have. Wang and Zhou~\cite{wang2007analyzing} demonstrated that the performance of co-training algorithms was largely affected by the complementary information in distinct views. These two principles are very important for multi-view learning and should be taken into consideration when designing multi-view learning algorithms.

Xu \emph{et al.} also categorized the classical approaches of combining multiple views into co-training style algorithms, multiple kernel learning algorithms, and subspace learning-based approaches~\cite{xu2013survey}.
\begin{itemize}
    \item \emph{Co-training style algorithm:} Blum and Mitchell \cite{blum1998combining} proposed the original co-training algorithm to solve semi-supervised classification problems. Firstly, two classifiers are trained separately on each view, and then each classifier labels the unlabeled data which are then added to the training set of another classifier. Besides, Kumar and Daum{\'e}~\cite{kumar2011co} extended the idea of co-training to an unsupervised setting. Since co-training algorithms usually separately train the base learners, it can be seen as a late combination method. 
    \item \emph{Multiple kernel learning algorithm:} Combining different kernels is another way to integrate multiple views and can be regarded as an intermediate method for the reason that kernels are integrated just before or during the training phase. These methods include linear combination methods~\cite{lanckriet2004learning, joachims2001composite} and nonlinear combination methods~\cite{cortes2009learning}.
    \item \emph{Subspace learning-based approach:} In the subspace learning-based approach, there is an assumption that multiple views are generated from a latent subspace. This approach can be regarded as a prior combination of multiple views and the goal is to obtain the latent subspace. Canonical correlation analysis (CCA)~\cite{hotelling1992relations} is a classical subspace learning-based approach that can be applied to the datasets that contain two views. Besides, it can be extended to cope with datasets represented by more than two views~\cite{kettenring1971canonical} and to kernel CCA~\cite{fyfe2000ica}.
\end{itemize}

\textbf{Federated Learning:} Federated Learning (FL) was proposed by McMahan \emph{et al.}~\cite{mcmahan2017communication}. It is a collaborative machine learning paradigm for training models based on locally stored data from multiple organizations in a privacy-preserving way. Yang~\emph{et al.} gave a comprehensive survey for FL~\cite{yang2019federated}, which introduced horizontal federated learning, vertical federated learning, and federated transfer learning. 
\begin{itemize}
    \item \emph{Horizontal Federated Learning:} In horizontal federated learning, or sample-based federated learning, the data sets share the same feature space but have different sample ID space~\cite{yang2019federated}. Smith \emph{et al.} proposed a novel framework for federated multi-task learning and considered high communication cost, stragglers, and fault tolerance in the federated environment for the first time~\cite{smith2017federated}. Bonawitz \emph{et al.} designed a secure aggregation scheme that allows a server to securely compute users' data from mobile devices~\cite{bonawitz2017practical}. 
    
    \item \emph{Vertical Federated Learning:} In vertical federated learning, or feature-based federated learning, the data sets share the same sample ID space but have different feature space~\cite{yang2019federated}. Some algorithms and models have been proposed for vertical federated learning.~\cite{ gascon2016secure, karr2009privacy, sanil2004privacy} are about secure linear regression. Du \emph{et al.} defined two types of secure 2-party multivariate statistical analysis problems: linear regression and classification problem. Secure methods also proposed to solve these two problems~\cite{du2004privacy}. Liu \emph{et al.} proposed asymmetrical vertical federated learning and showed the way to achieve the asymmetrical ID alignment~\cite{liu2020asymmetrically}.
    
    \item \emph{Federated Transfer Learning:} In federated transfer learning, the data sets are different in both samples and feature space~\cite{yang2019federated}. Liu \emph{et al.} proposed an end-to-end approach to the FTL problem and demonstrated it was comparable to transfer learning without privacy protection~\cite{liu2020secure}. A federated transfer learning framework for wearable healthcare was proposed in~\cite{chen2020fedhealth}.
\end{itemize}

\textbf{Federated Multi-View Learning:} Federated learning with multi-view data is currently less well explored in the literature. Adrian Flanagan \emph{et al.} integrated multi-view matrix factorization with a federated learning framework for personalized recommendations and introduced a solution to the cold-start problem~\cite{flanagan2020federated}. Huang \emph{et al.} proposed FL-MV-DSSM, which is a generic content-based federated multi-view framework for recommendation scenarios and can address the cold-start problem~\cite{huang2020federated}. Xu \emph{et al.}~\cite{xu2021federated} extend the DeepMood model~\cite{cao2017deepmood} to a multi-view federated learning framework that suits the horizontal case. Feng \emph{et al.} extended the idea of multi-view learning and proposed MMVFL that enables label sharing from its owner to other participants~\cite{feng2020multi}. MMVFL suits for vertical setting and it can deal with multi-participant and multi-class problems while the other existing VFL approaches can only handle two participants and binary classification problems. Kim \emph{et al.} proposed a federated tensor factorization framework for horizontally partitioned data~\cite{kim2017federated}. To sum up, today's federated multi-view learning is mainly applied to recommendation systems or tenor data, or only consider the horizontal or vertical situation, rather than consider the two together.

\section{Conclusion} \label{sec:conclusion}
In this paper, we proposed a generic multi-view learning framework by using federated learning paradigm for privacy-preserving and secure sharing of medical data among institutions, which can well protect the patient privacy by keeping the data within their own confines and achieve multi-view data integration.
Specifically, we investigated two types of multi-view learning (i.e., vertical and horizontal data integration) in the setting of federated learning based on different local data availability, and developed the vertical federated multi-view learning (V-FedMV) and horizontal federated multi-view learning (H-FedMV) algorithms  to solve this problem. Moreover, we adapted our model to deal with multi-view sequential data in a federated environment and introduced a federated multi-view sequential learning (S-FedMV) method. 
Extensive experiments on real-world keyboard data demonstrated that our methods can make full use of multi-view data and obtain better classification results as compared to local training. Moreover, the result of S-FedMV is comparable with the result of centralized method that cannot protect the privacy of sensitive data and thus showing the effectiveness of our federated method.

\section*{Acknowledgments}
This work is supported by NSF ONR N00014-18-1-2009.
Hao Peng is supported by NSFC program (No. 62002007 and U20B2053), and S\&T Program of Hebei through grant 20310101D.

\bibliographystyle{ACM-Reference-Format}
\bibliography{reference}

\end{document}